\documentclass[11pt]{article}
\usepackage{coling2018}
\usepackage{times}
\usepackage{url}
\usepackage{latexsym}
\usepackage{graphicx}

\title{Emotion Detection in Text: a Review}

\author{Armin Seyeditabari \\
  UNC Charlotte \\
   {\tt sseyedi1@uncc.edu} \\\And
  Narges Tabari \\
  UNC Charlotte \\
  {\tt nseyedit@uncc.edu} \\\And
Wlodek Zadrozny \\
UNC Charlotte \\
{\tt wzadrozn@uncc.edu} \\}
\date{}

\begin{document}
\maketitle
\begin{abstract}
In recent years, emotion detection in text has become more popular due to its vast potential applications in marketing, political science, psychology, human-computer interaction, artificial intelligence, etc. Access to huge amount of textual data, specially opinionated and self expression text also played a special role to bring attention to this field. In this paper, we review the work that has been done in identifying emotion expressions in text, and argue that although many techniques, methodologies and models have been created to detect emotion in text, there are various reasons that makes these methods insufficient. Although, there is an essential need to improve the design and architecture of current systems, factors such as the complexity of human emotions, and the use of implicit and metaphorical language in expressing it, lead us to think that just re-purposing standard methodologies will not be enough to capture these complexities, and it is important to pay attention to the linguistic intricacies of emotion expression.
\end{abstract}

\section{Introduction}
\label{intro}

%
%
\blfootnote{
    %
    %
    \hspace{-0.65cm}  
    Placeholder for licence statement.
    %
    %
    %
    %
}

Emotion detection in computational linguistics is the process of identifying discrete emotion expressed in text. Emotion analysis can be viewed as a natural evolution of sentiment analysis and its more fine-grained model. However, as we observe in this paper, this field still has a long way to go before matching the success and ubiquity of sentiment analysis. 

Sentiment analysis, with thousands of articles written about its methods and applications, is a well established field in natural language processing. It has proven very useful in several applications such as marketing, advertising \cite{Qiu2010,Jin:2007}, question answering systems \cite{somasundaran2007qa,stoyanov2005multi,lita2005qualitative}, summarization\cite{seki2005multi} , as part of recommendation systems \cite{Terveen:1997}, or even improving information extraction \cite{Riloff2005}, and many more. 

On the other hand, the amount of useful information which can be gained by moving past the negative and positive sentiments and towards identifying discrete emotions can help improve many applications mentioned above, and also open ways to new use cases. In other words, not all negative or positive sentiments are created equal. For example, the two emotions \textit{Fear} and \textit{Anger} both express negative opinion of a person toward something, but the latter is more relevant in marketing or socio-political monitoring of the public sentiment. It has been shown that fearful people tend to have pessimistic view of the future, while angry people tend to have more optimistic view \cite{lerner2000beyond}. Moreover, fear generally is a passive emotion, while anger is more likely to lead to action \cite{miller2009relative}. 

These more precise types of information on the nature of human emotions indicate potential uses of emotion detection. The usefulness of understanding emotions in political science \cite{Druckman2008}, psychology, marketing \cite{bagozzi1999role}, human-computer interaction \cite{brave2003emotion}, and many more, gave the field of emotion detection in natural language processing life of its own, resulting in a surge of research papers in recent years. In marketing, emotion detection can be used to analyze consumers reactions to products and services to decide which aspect of the product should be changed to create a better relationship with customers in order to increase customer satisfaction \cite{gupta2013emotion}. Also emotion detection can be used in human computer interaction and recommender systems to produce interactions or recommendations based on the emotional state of the user \cite{voeffray2011emotion}. Results of emotion detection systems can also be used as input to other systems, like what \newcite{rangel2016impact} has done in profiling authors by analyzing the presence of emotions in their text. By understanding the important role of emotions in decision making process in humans \cite{BECHARA200430}, emotion detection can profit any entity or organization that wants to assess the impact of their products and actions on the population, and to be able to manage people's reactions by monitoring their emotional responses. Thus understanding emotions can benefit any entity and organization such as commercial institutes, political campaigns, managing the response to a natural disaster. One can also argue it is necessary to create better artificial intelligence tools, e.g. chatbots.

The main contribution of this paper to the computational linguistic community is to provide a review of the literature in this field, and to summarize the work that has already been done, the shortcomings, and the avenues to move the field forward. 
Reviewing the literature shows that identifying emotions is a  hard task. It is mainly because of two factors, firstly emotion detection is a multi-class classification task combining multiple problems of machine learning and natural language processing; and the second is the elusive nature of emotion expression in text, which comes from the complex nature of the emotional language (e.g. implicit expression of emotions, metaphors, etc.), and also the complexity of human emotions. We believe this review fills a significant gap, since, to the best of our knowledge, there has not been a comprehensive review paper focused specifically on emotion detection in text, and arguably the topic is important.

In this paper first talk about models and theories of emotions in psychology, to quickly get an idea about what models of emotions are, How they have been categorized in a discrete or continuous space is the topic of Section 2. Then we will focus on the reasons behind the linguistic complexity of this task in Section 3. In the subsequent three sections, we review the resources and methodologies used for detecting emotion in text. The current state of the field and future work is discussed in Section 7. And finally, we conclude our work in Section 8.

\section{Psychological Models of Emotion}

The prerequisite for talking about extracting emotions, is having a general idea about the emotion models and theories in psychology. This body of research provides us with definitions, terminology, models, and theories. Here we introduce the most general and well accepted theories in a short section
to give the reader the basic information needed for the rest of the paper. 

In psychology, and based on the appraisal theory, emotions are viewed as \textit{states that reflect evaluative judgments (appraisal) of the environment, the self and other social agents, in light of the organism’s goals and beliefs, which motivate and coordinate adaptive behavior} \cite{hudlicka2011guidelines}. In psychology, emotions are categorized into basic emotions, and complex emotions (i.e. emotions that are hard to classify under single term such as guilt, pride, shame, etc.). In this paper when we talk about emotions, we mostly mean basic emotions.

 Although there is no universally accepted model of emotions, some of the most widely accepted models that have been used in emotion detection literature can be divided based on two viewpoints: emotions as discrete categories, and dimensional models of emotions. According to Discrete Emotion Theory, some emotions are distinguishable on the basis of neural, physiological, behavioral and expressive features regardless of culture \cite{colombetti2009affect}. A well known and most used example is Ekman's six basic emotions \cite{ekman1992argument}. Ekman et al. in a cross-cultural study found six basic emotions of \textit{sadness, happiness, anger, fear, disgust, and surprise}. Most papers in emotion detection used this model for detecting emotions as a multi-class classification problem, along with some that are based on Plutchik's wheel of emotions \cite{plutchik1984emotions} in which he categorized eight basic emotions (\textit{joy, trust, fear, surprise, sadness, disgust, anger and anticipation}) as pairs of opposite emotions. \newcite{parrott2001emotions}, in his three layered categorization of emotion, considered six primary emotions of \textit{love. joy, surprise, anger, sadness, and fear} in the first layer, followed by 25 secondary emotions in the next. He categorized more fine grained emotions in the last layer.



Using a different perspective, dimensional model of emotions tries to define emotions based on two or three dimensions. As opposed to basic emotions theory, which states that different emotions correspond to different neurological subsystems in the brain, the dimensional model is based on the hypothesis that all emotions are the result of a common and interconnected neurophysiological system. The Circumplex model developed by \newcite{russell1980circumplex} suggests that emotions can be shown in a two dimensional circular space, with one dimension for arousal (i.e. intensity), and one for valance (i.e. pleasantness). The dimensional models have  been used very scarcely in the emotion detection literature, but shown to be promising as a model to represent emotions in textual data \cite{calvo2013emotions}.
%

\section{Complexity of Expressing Emotions in Language}
Emotion expression is very context sensitive and complex. \newcite{ben2000subtlety} relates this complexity to various reasons: first, its sensitivity to multiple personal and contextual circumstances; secondly, to the fact that these expressions often consist of a cluster of emotions rather than merely a single one; and finally, the confusing linguistic use of emotional terms. \newcite{bazzanella2004emotions} argues that complexity of emotions can be seen in multiple levels: "the nested interplay with mind/language/behavior/culture, the lexical and semantic problem, the number of correlated physiological and neurological features, their universality or relativity, etc.". As one can see even in everyday life, it is sometimes very hard to distinguish between emotions. 

Also, it has been shown that context is very important, and is crucial in understanding emotions \cite{oatley2006understanding}. Most recent studies in textual emotion detection in NLP, are based on explicit expression of emotion using emotion bearing words. But emotion expression is mostly done by expressing emotion provoking situation, which can be interpreted in an affective manner \cite{balahur2008applying,pavlenko2008emotion}. This fact has greatly limited the identification of emotions, for considerable portion of these expressions are not explicit. Therefore more emphasis should be placed on implicit expressions of emotions \cite{lee2015linguistic}. 

There are not many works in the literature on detecting implicit expression of emotions, but in sentiment analysis literature there has been some attempts in this area. For instance,  \newcite{greene2009more} used syntactic packaging for ideas to assess the implicit sentiment in text, and to improve state of the art sentiment detection techniques. \newcite{cambria2009affectivespace} proposed an approach to overcome this issue by building a knowledge base that merges Common Sense and affective knowledge. The goal is to move past the methods that rely on explicit expression of emotion i.e. verbs, adjectives and adverbs of emotion. Their reasoning for choosing this approach was based on the notion that most emotions are expressed through concepts with affective valence. For example 'be laid off' or 'go on a first date' which contains emotional information without specifying any emotional lexicon.

\newcite{lakoff2008women},  in a case study about \textit{Anger}, talks about conceptual content behind emotions. He argues that emotions have a very complex conceptual structure, and this structure could be studied by systematic investigation of expression that are understood metaphorically. He argues that many expressions of anger are metaphorical, thus could not be assessed by the literal meaning of the expression (e.g. 'he lost his cool' or 'you make my blood boil'). This fact makes it more difficult to create a lexical, or machine learning method to identify emotions in text, without first solving the problem of understanding of metaphorical expressions.

Complexity of human emotions, along with implicit expressions, frequent use of metaphors, and the importance of context in identifying emotions, not to mention cross cultural and intra-cultural variations of emotions, rises the problem of detecting emotions from text above a multi-class classification problem which covers the most research that has been done in the field.

\section{Resources for Detecting Emotions in Text}

As opposed to sentiment analysis, textual datasets annotated with markers of emotional content are scarce. Any new method of emotion detection in text, based on conventional supervised classifiers or neural networks, requires vast amount of annotated data for training and development. But as a relatively new field in natural language processing, emotion detection as a multi-class classification problem, faces lack of available annotated data. In this section, some of the most prominent and publicly available sources will be introduced. These data can be separated into two groups: labeled texts and emotion lexicons. We will also briefly cover vector space models, as another potential resource.  

\subsection{Labeled Text}

Having a standard, free and generalized annotated data makes it easier to train and test any new method, and is an important factor in any classification task. One of the most prominent and well known sources for emotionally labeled text is the Swiss Center for Affective Sciences \cite{SCASweb}. The most used resource they provide is ISEAR, International Survey On Emotion Antecedents And Reactions. It consists of responses from about 3000 people around the world who were asked to report situations in which they experienced each of the seven major emotions (joy, fear, anger, sadness, disgust, shame, and guilt), and how they reacted to them. The result was a promising dataset to be used to test many methods for emotion extraction and classification. This dataset consists of about 7600 records of emotion provoking text. SCAS has many more resources that can be useful specially in languages other than English.

EmotiNet knowledge base \cite{Balahur2011} tackled the problem of emotion detection from another perspective. Balahur et. al. argued that word level attempt to detect emotion would lead to a low performance system because "expressions of emotions are most of the time not presented in text in specific words", rather from the "interpretation of the situation presented" in the text. They base their insight on \textit{Appraisal Theory} in psychology \cite{dalgleish2000handbook} . They create a new knowledge base containing action chains and their corresponding emotional label. They started from around a thousand samples from the ISEAR database and clustered the examples within each emotion category based on language similarity. Then they extracted the agent, the verb and the object from a selected subset of examples. Furthermore they expanded the ontology created using VerbOcean \cite{chklovski2004verbocean} in order to increase the number of actions in the knowledge base, and to reduce the degree of dependency between the resources and the initial set of examples. Although this approach showed  promise, specially because of their attempt to extract concepts from text, it could not present itself as viable and generally applicable in its current form, due to the small size of the knowledge base and the structure of information they used (limited to the four-tuple of actor, action, object, and emotion). 

\newcite{vu2014acquiring} focused on discovery and aggregation of emotion provoking events. They created dictionary of such events through a survey of 30 subjects, and used that to aggregate similar events from the web by applying Espresso pattern expansion \cite{Pantel:2006} and bootstrapping algorithms. One of the frequently used dataset is the SemEval-2007 \cite{Strapparava:2007:STA:1621474.1621487}, which consists of 1250 news headlines extracted from news websites, and annotated with six Ekman's emotions. The other example, is Alm's annotated fairy tale dataset \cite{alm2005emotions}, consisting of 1580 sentences from children fairy tales, also annotated with six Ekman's emotions. These datasets have been mostly used as benchmark in the literature. As emotion detection gets more attention, there will be the need for more datasets that could be used in different tests of models and methods for emotion detection. 

In the meantime the lack of benchmark datasets with proper linguistic generality and accepted annotations pushes the research community to use text from microblogs, such as Twitter, in which self expression is possible using methods like \textit{hashtags, and emoticons}. An attempt to create such an annotated corpus was presented in \newcite{wang2012harnessing} consisting of 2.5 million tweets cleaned and annotated with hashtags and emoticons.

\subsection{Emotion Lexicons}

Although having expressive emotional text like ISEAR is very important, especially for comparing different emotion detection methods, there are many use cases in which having an annotated lexicon could be useful, specially when more word based analysis is required. And even though considerable data is available on sentiment polarity of words going back a few years \cite{baccianella2010sentiwordnet}, the lack of reasonable size lexicon for emotions led \newcite{Mohammad:2010:EEC:1860631.1860635} to create an emotion word lexicon. In the cited paper and later in \cite{Mohammad13} they used Amazon Mechanical Turk to annotate  around 14000 words in English language (along with lexicons in other languages, these are available on their website\footnote{NRC word-emotion association lexicon: http://saifmohammad.com/WebPages/NRC-Emotion-Lexicon.htm}). 

Another popular emotion lexicon used in literature is WordNet-Affect.  \newcite{strapparava2004wordnet} tried to create a lexical representation of affective knowledge by starting from WordNet \cite{miller1998wordnet}, a well known lexical database. Then they used selection and tagging of a subset of synsets which represents the affective concepts, with the goal of introducing "affective domain labels" to the hierarchical structure of WordNet. WordNet-Affect, despite its small size (containing 2874 synsets, and 4787 words), was a great attempt to extract emotional relations of words from WordNet, and was used in many early applications of sentiment analysis, opinion mining \cite{balahur2013sentiment}, and in emotion detection specially for extending affective word sets from the basic set of emotions.

Another attempt to generate an emotional lexicon has been showcased by \newcite{staiano2014Depeche} called DepecheMood. They used crowd-sourcing to annotate thirty five thousands words. The showed that lexicons, could be used in several approaches in sentiment analysis, as features for classification in machine learning methods \cite{liu2012survey}, or to generate an affect score for each sentence, based on the scores of the words which are higher in the parse tree \cite{socher2013recursive}. Other emotional lexicons frequently used in the literature are LIWC lexicon \cite{pennebaker2001linguistic} consisting 6400 words annotated for emotions, and also ANEW (Affective Norm for English Words) developed by \newcite{bradley1999affective}. This dataset has near 2000 words which has been annotated based on dimensional model of emotions, with three dimensions of valance, arousal and dominance.

\subsection{Word Embedding}

Word embeddings is a technique based on distributional semantic modeling. It is rooted in the idea that words which frequently co-occur in a relatively large corpus are similar in some semantic criteria. In these methods, each word is represented as a vector in an n-dimensional space, called the vector space, and in a way that the distance between vectors corresponds to the semantic similarity of the words they represent. These vector space models have been shown to be useful in many natural language processing tasks, such as named entity recognition \cite{tirian2010}, machine translation \cite{zou2013bilingual}, and parsing \cite{socher2013parsing}. Many such models have been created in recent years with similar performances as shown by \newcite{levy2015improving}. Some of the more well-established and most frequently used embedding models in the literature are latent semantic analysis or LSA, Word2Vec \cite{mikolov2013efficient,mikolov2013distributed}, GloVe \cite{pennington2014glove}, and ConceptNet \cite{speer2016conceptnet}. It has been shown that these models, just by utilizing the statistical information of word co-occurrences, can incorporate variety of information about words \cite{pennington2014glove} such as closeness in meaning, gender, types, capital of countries, etc., and in the arithmetic of word vectors shown in such overused examples as $v(king) - v(queen) = v(man) - v(woman)$.

There also have been many attempts to increase their performance, and incorporate more information in these models retrofitting \cite{faruqui2014retrofitting} and counter-fitting \cite{mrkvsic2016counter}  external word ontologies or lexicons \cite{speer2016conceptnet,speer2016ensemble}. Some work has been done in creating embeddings for sentiment analysis. For example, by \newcite{tang2014learning} who created a sentiment-specific word embeddings using neural networks, to classify sentiments in Twitter \cite{tang2014coooolll}. Such approaches for creating emotional word embeddings from scratch, or incorporating emotional information into pre-trained word vectors after the fact, might lead to better performances in emotion detection tasks, either in unsupervised methods, or as features for classification tasks using conventional machine learning, or deep learning \cite{socher2013recursive}.


\section{Methodologies for Detecting Emotions in Text: Supervised Approaches}

Due to the lack of emotion-labeled datasets, many supervised classifications for emotions have been done on data gathered from microblogs (e.g. Twitter), using hashtags or emoticons as the emotional label for the data, under the assumption that these signals show the emotional state of the writer. Such an attempt can be seen in \newcite{suttles2013distant}, where the four pairs of opposite emotions in the Plutchik's wheel were used to create four binary classification tasks. With hashtags, emoticons, and emoji as labels for their data, they reached between 75\% to 91\%  accuracy on a separate manually labeled dataset. 

\newcite{purver2012experimenting} on Twitter data using SVM classifier reached 82\% accuracy for classifying the emotion \textit{Happy} in 10-fold cross validation, and 67\% in classifying over the entire dataset for the same emotion, with emoticons as labels for the training set, and hashtags as labels for the test set. Then they tested their trained models for each emotion to see if they can distinguish emotion classes from each other rather than just distinguish one class from a general \textit{Other} set.  The results varied from 13\% to 76\% accuracy for different emotions. They also created a dataset of 1000 tweets labeled by human annotators 
and used it as the test data to evaluate the quality of assigning hashtags and emoticons as labels. For different emotions the F-score varied from 0.10 to 0.77. Their study showed that the classifiers performed well on emotions like happiness, sadness and anger, but not well for others. They concluded that using hashtags and emoticons as labels is a promising labeling strategy and and alternative to manual labeling.

\newcite{mohammad2012emotional} also used hashtags as label for tweets, and used support vector machines as a binary classifier for each emotion in Ekman's model. After showing that the hashtags as labels perform better than random classification, he used Daumé's domain adaptation method \cite{daume2009frustratingly} to test the classification power of their data in a new domain.
\newcite{roberts2012empatweet} collected tweets in 14 topics that "would frequently evoke emotion" and created a dataset where all seven emotions (Ekman + Love) were represented. 
Seven SVM binary classifiers were used to detect emotions in the dataset, resulting in the average F1-score of 0.66.  

\newcite{hasan2014emotex} also used hashtags as their labels and created their features using the unigram model, removing any word from tweets which were not in their emotion lexicon (created using 28 basic emotion word in Circumplex model and extended with  WordNet synsets). Four classifiers (Naive Bayes, SVM, Decision Tree, and KNN) achieved accuracies close to 90\% in classifying four main classes of emotion categories in Circumplex model. In another paper, \newcite{Hasan2018} created an automatic emotion detection system to identify emotions in streams of tweets. This approach included two tasks: training an offline emotion classification model based on their 2014 paper, and in the second part a two step classification to identify tweets containing emotions, and to classify these emotional tweets into more fine-grained labels using soft classification techniques.

Facing the problem of lack of labeled emotional text, \newcite{wang2012harnessing} created a large dataset (about 2.5 million tweets) using emotion related hashtags, and used two machine learning algorithms for emotion identification. They used \newcite{shaver1987emotion} for mapping hashtags to emotions, and extending hashtag words to the total of 131 for the seven basic emotions. They then increased the quality of the data by keeping more relevant tweets (i.e. tweets with hashtags at the end of sentence, with more than 5 words, contain no URLs or quotations, in English, and containing less than 4 hashtags), and tried different combinations of features (e.g. different n-grams, position of n-grams, multiple lexicon, POS) with 250k of the training data to find the best set of features, with the best result for the combination of n-gram(n=1,2), LIWC lexicon, MPQA lexicon, WordNet-Affect, and POS. After finding the best feature set, they increased the size of training data from 1000 tweets to full training set to see the effect of training size in the classification. The final classifier reached the F-Measure as high as 0.72 for joy, and as low as 0.13 for surprise. They justified the varying result for different emotions by the fact that the training dataset had and unbalanced distribution. In addition, based on the confusion matrix, they reported that high number of misclassified tweets between class pairs like anger and sadness, or joy and love, were due to the fact that these emotions are "naturally related", and "different people might have different emotions when facing similar events." 

%

In another Twitter emotion classification task done by \newcite{balabantaray2012multi}, manual labeling was used for around 8000 tweets, for six basic emotions in Ekman's model. They used SVM multi-class classifier with 11 features: \textit{Unigrams, Bigrams, Personal-pronouns, Adjectives, Word-net Affect lexicon, Word-net Affect lexicon with left/right context, Word-net Affect emotion POS, POS, POS-bigrams, Dependency-Parsing, and Emoticons} resulting in an accuracy of 73.24\%.


We can see combination of methods in emotion classification in the paper by \newcite{wen2014emotion}. In their study, they used a combination of lexicon based and machine learning (SVM) methods to create two emotion labels for each microblog post, they then use Class Sequential Rules (CSR) \cite{liu2007web} mining to create sequences for each post based on the labeling for each sentence and the conjunctions between them. Using the resulting data and by including additional features like lexicon counts and punctuations, and using an SVM classifier they reached an F-measure of 0.44 which was shown to be a significant increase over other methods based on emotion lexicons or simple SVM.

%
%

\newcite{li2014text} proposed a "emotion cause detection technique" to extract features that are "meaningful" to emotions instead of choosing words with high co-occurrence degree. Their method is based on Lee et al.'s work on rule based emotion cause detection \cite{lee2010text}. After using predefined linguistic patterns to extract emotion causes and adding it to their features, they used Support Vector Regression (SVR) to create the classifier, and reached higher F-score for some emotions like happiness, anger, and disgust compared to previous works. Overall, their approach had better precisions, but low recalls.

In their paper, \newcite{li2015sentence} attempted sentence level classification of emotion instead of document level. They indicated that the two biggest problems in sentence level emotion classification is firstly the fact that it is a multi-class classification, meaning that each sentence could have more than one label, and secondly, the short length of a sentence, provides less content. Considering these challenges they created a Dependence Factor Graph (DFG) based on two observations, \textit{label dependence}, i.e. multiple labels for a sentence would be correlated to one another, like Joy and Love instead of Joy and Hate, and \textit{context dependence}, i.e. two neighboring sentences, or sentences in the same paragraph might share the same emotion categories. Using the DFG model, after learning they reached the accuracy of 63.4\% with F1 of 0.37 showing significant improvement over previous methods \cite{wang2014enhancing,xu2012coarse}.

In an application based study done by \newcite{seyeditabari2018cross}, they attempted to classify social media comments regarding a specific crisis event, based on the emotion of anger considering the fact that the same method can be use for other emotions. They ran a short survey gathering 1192 responses in which the participants were asked to comment under a news headline as though they are commenting on social media. Using this as the training set they reached 90\% accuracy in classifying anger in a dataset created using the same survey from different population by using logistic regression coefficients to select features (words) and random forest as the main classifier. 

Current state of the art algorithms for emotion classification, are mostly based on supervised methods, but imbalance training data, specially for emotion detection as a multi-class classification problem, are an obstacle for supervised learning, leading to increase misclassification for underrepresented classes \cite{lopez2013insight,yang200610,wang2012harnessing}. There are different methods proposed in literature \cite{lopez2013insight} to overcome this issue in one of three ways, either by changing the learning algorithm to adapt to this imbalance \cite{tang2009svms}, or adding cost to majority classes during training \cite{sun2007cost}, or by sampling from the training data before learning to make the classes balanced \cite{chawla2004special}. \newcite{xu2015word} proposed an over-sampling method based on word embeddings \cite{mikolov2013distributed}, and recursive neural tensor network \cite{socher2013recursive} which showed a significant improvement over previous sampling methods, specially for emotion classification as a multi-class data.

The question here could be if creating emotion detection systems based on conventional machine learning techniques can move past the mediocre results we have seen in the literature. To emphasis the importance of a deeper analysis than conventional machine learning methods we can refer to a comparative analysis done by \newcite{balahur2012detecting}. They compared various classification features and compared them to EmotiNet, and concluded that the task of emotion detection can be best tackled using approaches based on commonsense knowledge. They showed that even with the small size of EmotiNet knowledge base they could produce comparative results to supervised learning methods with huge amount of training data.

\section{Methodologies for Detecting Emotions in Text: Unsupervised Approaches}

 \newcite{Kim2010evaluation} used an unsupervised method to automatically detect emotions in text, based on both categorical (anger, fear, joy and sadness), and dimensional models of emotions. They used three datasets, SemEval-2007 “Affective Text”, ISEAR, and children’s fairy tales. For categorical model, they used WordNet-Affect as the lexicon, and evaluated three dimensionality reduction methods: Latent Semantic Analysis (LSA), Probabilistic Latent Semantic Analysis (PLSA), and Non-negative Matrix Factorization (NMF). And for the dimensional model, they used ANEW (Affective Norm for English Words) and WordNet-Affect as a means to extend ANEW. They assigned the emotion of the text based on closeness (cosine similarity) of its vector to the vectors for each category or dimension of emotions. Their study showed that NMF-based categorical classification performs best among categorical approaches, and dimensional model had the second best performance with highest F-measure of 0.73.

Another unsupervised approach to emotion detection can be seen in the paper by  \newcite{agrawal2012unsupervised}. They start by extracting NAVA words (i.e. Nouns, Adjectives, Verbs, and Adverbs) from a sentence, and then extracting syntactic dependencies between extracted words in each sentence to include contextual information in their model. They then used semantic relatedness to compute emotion vectors for words, based on the assumption that the affect words (NAVA words) which co-occur together more often tend to be semantically related. They use Point-wise Mutual Information (PMI) as the measure of semantic relatedness of two words (Equation \ref{eq:pmi}) and computed a vector for each word using the PMI of the word with all words related to each emotion, then adjust the vectors by considering the contextual information in syntactic dependency of words. After computing vectors for each word they generated a vector for each sentence by aggregating the emotion vectors of all the affect words in it. By evaluating on multiple data sources, they showed that their method preformed more accurate, compared to other unsupervised approaches, and had comparable results to some supervised methods.

\begin{equation} \label{eq:pmi}
PMI(x,y)= \frac{cooccurrence(x,y)} {occurrence(x) * occurrence(y)}
\end{equation}

 In another lexicon base approach, \newcite{mohammad2012once} showed how detecting emotions in text can be used to organize collection of text for affect-based search, and how books portray different entities through co-occurring emotion words by analyzing emails, and books. He used NRC lexicon to see which of the emotion words exist in the available text, and calculated ratios such as the number of words associated to a particular emotions compared to other emotions, to determine if a document have more expressed emotions compared to other documents in the corpus. He compiled three datasets for emotional emails: \textit{love letters, hate mails, and suicide notes}. He goes on to analyze  presents of different emotions based on criteria like, workplace emails, emails written by women/men, or emails written by men to women vs men. He also did some fascinating analysis on books, and works of literature using the same lexical approach.

%
%

\newcite{rey2016analysis} used an unsupervised method to distinguish language pattern related to anxiety in online health forums. They define user behavioral dimension (BD) based on the LIWC lexicon focusing on its anxious word list. They define each user's BD as measure of the average fraction of words from the list $BD_{i}$ across the posts of a user, Equation \ref{eq:BD}:

\begin{equation} \label{eq:BD}
BD_i(u) = log\bigg(\frac{1}{|posts(u)|} \sum_{p \in posts(u)}^{} \frac{|words_{BD_i}(p)|}{|words(p)|}\bigg)
\end{equation}

Then by analyzing this value for each user or groups of users over time, and the correlation of this behavioral dimension with other BDs, they showed that the anxiety level of patients involved in a support group lowers over time. In a rule based approach \newcite{tromp2014rule} introduced RBEM-Emo method for detecting emotions in text as an extension of their previous work for polarity detection \cite{tromp2013rbem} called Rule-Based Emission Model. They showed that rule based classification techniques can be comparative to current state of the art machine learning methods, such as SVM classifier and recursive auto-encoder.

 \newcite{BANDHAKAVI2017133} used domain-specific lexicon that they created based on unigram mixture models \cite{bandhakavi2014generating,bandhakavi2017lexicon} to extract features and showed that their lexicon outperform methods like Point-wise Mutual Information, and supervised Latent Dirichlet Allocation.

\section{Discussion and Open Problems}

Going through the literature, we can see the hard task of detecting expressed emotions. The difficulties can be attributed to many problems from \textit{complex nature of emotion expression in text}, to \textit{inefficiency of current emotion detection models}, and \textit{lack of high quality data} to be utilized by those models. 

{\bf Complex Nature of Emotion Expression}: On one hand, expression of emotion in human is a complex phenomena, in such a way that a shortest phrase can express multiple emotions with different intensity that cannot be understood at first glance even by humans. And on the other hand, the intricacy of emotional language, resulting from the vast use of metaphorical language, context dependent nature of emotion expression, and implicit nature of such expressions, makes this task even harder. In order to address this issue, it is important to pay attention to the complexity of emotional language when building emotion detection systems. These systems should be designed based on the linguistic complexities of emotion expression to be able to grasp the implicit expression of emotions, and untangle the metaphorical nature of these expressions. It is also crucial to consider the contextual information in which the expression is occurring.

 {\bf Shortage of Quality Data}: In almost all the papers reviewed, some common obstacles can be identified, showing that future work is needed in order to improve performance of emotion detecting systems. In any machine learning task, the quality and quantity of data has a huge effect on the performance of classification algorithms. Although huge amount of textual data is currently available, for any supervised model, a large amount of annotated data is required. A great body of work has already been dedicated to overcome this problem by using self annotated microblog data, but it has not yet possesses qualities which are required for an applicable system. Additionally, the niche nature of the language used in microblog text, prevents the systems trained on these texts to be used to classify other types of text (e.g. tweets vs. news comments). Furthermore, as can be seen in most of the reviewed studies, the imbalance nature of currently available emotional text, will cause the classifier to severely under-preform for emotions that are underrepresented in the dataset. Therefore, any attempt to create a large balanced dataset, with high quality labels could provide a brighter future for the field. 

{\bf Inefficiency of Current Models}: In addition, creating a multi-class classification methodology based on the nature of the data and the task at hand, is another front that could be considered to increase the performance of such systems. There have been many attempt to approach this problem with the most frequently used being, converting the task of multi-class to multiple binary classification, either by having one classifier for each emotion (e.g. anger vs not anger), or one classifier for a pair of opposite emotions (e.g. joy vs sadness). Further improvement in classification algorithms, and trying out new ways is necessary in order to improve the performance of emotion detection methods. Some suggestions that were less present in the literature, are to develop methods that go above BOW representations and consider the flow and composition of language. In addition, specific neural network designs or ensemble methods are possible approaches that has been shown to be useful in other areas of natural language processing. New ways to increase the emotional qualities of embeddings and vector models could be beneficial in unsupervised methods, or be used as features in neural networks. Emotion detection, as a lesser known and relatively new field, has come a long way, but still has a long way to go to become a totally reliable and applicable tool in natural language processing.

\section{Conclusion}

In this paper, we reviewed the current state of emotion detection in textual data based on the available work in the literature. While many successful methodology and resources was introduces for sentiment analysis in recent years, researchers, by understanding the importance of more fine-grained affective information in decision making, turned to emotion detection in order to distinguish between different negative or positive emotions. In addition, having large amount of textual data with the rise of social media in past couple of decades, and therefore the availability of vast self expression text about any major or minor event, idea, or product, points to a great potential to change how entities and organizations can use these information as a basis for their future decision making processes.
%

\bibliographystyle{acl}
\bibliography{qual}

%

\end{document}